\documentclass[conference]{IEEEtran}
\IEEEoverridecommandlockouts
\usepackage{cite}
\usepackage{amsmath,amssymb,amsfonts}
\usepackage{algorithmic}
\usepackage{graphicx}
\usepackage{textcomp}
\usepackage{xcolor}
\usepackage{enumitem}

\usepackage{multirow}
\usepackage{xcolor}
\usepackage{siunitx}
\def\BibTeX{{\rm B\kern-.05em{\sc i\kern-.025em b}\kern-.08em
    T\kern-.1667em\lower.7ex\hbox{E}\kern-.125emX}}
\begin{document}

\title{A Multi-Component AI Framework for Computational Psychology: From Robust Predictive Modeling to Deployed Generative Dialogue}

\author{
    \IEEEauthorblockN{Anant Pareek}
    \IEEEauthorblockA{\textit{Independent Researcher} \\
    Darjeeling, India \\
    apanantpareek@gmail.com \\
    ORCID: 0009-0004-5967-2896}

}

\maketitle

\begin{abstract}The confluence of Artificial Intelligence and Computational Psychology presents an opportunity to model, understand, and interact with complex human psychological states through computational means. This paper presents a comprehensive, multi-faceted framework designed to bridge the gap between isolated predictive modeling and an interactive system for psychological analysis. The methodology encompasses a rigorous, end-to-end development lifecycle. First, foundational performance benchmarks were established on four diverse psychological datasets using classical machine learning techniques. Second, state-of-the-art transformer models were fine-tuned, a process that necessitated the development of effective solutions to overcome critical engineering challenges, including the resolution of numerical instability in regression tasks and the creation of a systematic workflow for conducting large-scale training under severe resource constraints. Third, a generative large language model (LLM) was fine-tuned using parameter-efficient techniques to function as an interactive "Personality Brain." Finally, the entire suite of predictive and generative models was architected and deployed as a robust, scalable microservices ecosystem. Key findings include the successful stabilization of transformer-based regression models for affective computing, showing meaningful predictive performance where standard approaches failed, and the development of a replicable methodology for democratizing large-scale AI research. The significance of this work lies in its holistic approach, demonstrating a complete research-to-deployment pipeline that integrates predictive analysis with generative dialogue, thereby providing a practical model for future research in computational psychology and human-AI interaction.
\end{abstract}

\begin{IEEEkeywords}
Computational Psychology, Transformer Models, Large Language Models (LLMs), Emotion Recognition, Personality Prediction
\end{IEEEkeywords}
\section{Introduction}

The intersection of Artificial Intelligence and Computational Psychology has paved the way for a new frontier in understanding the human mind. AI algorithms, particularly those in Natural Language Processing (NLP), offer powerful tools for modeling the complex cognitive and affective processes underlying human language\cite{b1}. This has spurred advancements in specific domains such as emotion recognition, personality trait prediction, and the development of empathetic dialogue systems\cite{b2}. While these applications have demonstrated considerable promise, a significant challenge remains: integrating these disparate analytical capabilities into a single, cohesive system that can provide a holistic psychological "snapshot" of an individual.

This research addresses the critical gap between developing isolated, task-specific models and engineering a functional, deployed system capable of both multifaceted analysis and dynamic interaction. The work confronts not only the scientific challenge of modeling diverse psychological constructs but also the engineering hurdles of training, managing, and serving multiple large-scale AI models. Many academic endeavors focus on achieving a state-of-the-art result on a single benchmark, often overlooking the complexities of system integration and deployment. This project adopts a "full-stack" research paradigm, viewing the entire lifecycle, from data curation and model engineering, to generative AI fine-tuning and deployment as an integrated research problem. This perspective is essential for translating theoretical AI advancements into tangible tools for fields like human-computer interaction (HCI) and mental health technology\cite{b1}.

This work had four primary goals:
\begin{enumerate}
    \item To benchmark predictive models across a diverse range of psychological constructs, including Big Five personality traits and dimensional emotions, using four distinct datasets.
    \item To investigate and resolve critical engineering problems that arise when applying state-of-the-art transformer architectures to these tasks, particularly numerical instability in regression and severe computational resource constraints.
    \item To extend the system's capabilities beyond prediction by fine-tuning a generative large language model (LLM) to serve as an interactive, personality-aware dialogue agent.
    \item To architect and deploy the entire suite of models as a scalable application using a microservices architecture, demonstrating a complete research-to-deployment pipeline.
\end{enumerate}

This paper is structured to follow this research lifecycle. Section~II details the foundational work of data curation and baseline analysis. Section~III presents the core research challenges and contributions in advanced predictive modeling. Section~IV describes the methodology for fine-tuning the generative LLM. Section~V outlines the design of the microservices-based system architecture. Finally, Section~VI summarizes the project's contributions and proposes directions for future work.

\section{Foundational Analysis and Baselines}
A structured and thorough approach forms the backbone of any advanced AI engineering effort. This initial phase focused on three key areas: curating a diverse set of data sources to ensure model robustness, conducting a thorough exploratory analysis to guide modeling strategy, and establishing quantitative performance baselines to ground all subsequent experimentation in a clear, measurable context.

\subsection{Data Curation and Heterogeneity}

To build a system capable of modeling a wide spectrum of psychological phenomena, four distinct datasets were strategically selected. Their heterogeneity in terms of task, text length, data volume, and annotation schema was a deliberate choice designed to challenge the models and prevent overfitting to a single data modality.

\begin{description}[font=\bfseries, itemsep=1ex, leftmargin=0.5cm]
    \item[Essaysbig5:] This dataset consists of 2,468 essays annotated with the Big Five personality traits (Openness, Conscientiousness, Extraversion, Agreeableness, Neuroticism)\cite{b3} on a binary classification basis (high/low)\cite{b4}. Its primary challenge lies in its long-form text and relatively small sample size, demanding models that can capture deep contextual cues without overfitting\cite{b5}.

    \item[GoEmotions:] A large-scale dataset of 58,000 Reddit comments, human-annotated for 27 fine-grained emotion categories plus a 'Neutral' class\cite{b5}. Sourced from conversational online text, it is characterized by short, often noisy inputs and a complex multi-label structure, where a single comment can express multiple emotions simultaneously. This dataset tests a model's ability to handle scale, noise, and multi-label classification effectively.

    \item[PANDORA:] This is the first large-scale Reddit dataset annotated with personality and demographic data, containing over 17 million comments from more than 10,000 users\cite{b7}. For this project, the subset of 1,600 users labeled with the Big Five personality model was used. Unlike Essaysbig5, PANDORA frames the task as a regression problem, predicting continuous scores for each of the five traits, which presents a more nuanced and difficult modeling challenge.

    \item[EmoBank:] A corpus of 10,000 English sentences from diverse genres, annotated with dimensional emotion metadata according to the Valence-Arousal-Dominance (VAD) psychological model\cite{b8},\cite{b9}. This dataset poses a regression task to predict continuous values for Valence (the positivity/negativity of an emotion) and Arousal (the intensity of the emotion), grounding the analysis in a well-established theoretical framework of affect.
    
\end{description}
\subsection{Universal Preprocessing Pipeline and EDA}
To ensure consistency across these varied sources, a universal preprocessing pipeline was engineered. This pipeline standardized the text data by applying a sequence of cleaning operations, including lowercasing, removal of URLs, special characters, and extraneous whitespace. This step was crucial to create a uniform data foundation, ensuring that subsequent model performance differences could be attributed to architectural or methodological choices rather than artifacts of the source data.

A thorough Exploratory Data Analysis (EDA) was then conducted on the preprocessed datasets. This analysis yielded critical insights that informed the project's entire strategic direction. The EDA confirmed that Essaysbig5, with its long texts but few samples, would be highly susceptible to overfitting with large transformer models. The massive scale, short text length, and severe class imbalance of GoEmotions highlighted the need for efficient data handling and robust evaluation metrics. Finally, the continuous, real-valued distributions of the target variables in Pandora and EmoBank confirmed their status as challenging regression tasks, flagging the potential for issues that do not typically arise in classification settings.

\subsection{Baseline Modeling and Performance Benchmarks}
Before proceeding to complex deep learning models, it is essential to establish strong performance baselines. This practice provides an important sanity check and a quantitative measure against which more advanced models must demonstrate significant improvement to justify their computational expense. A suite of classical machine learning models were trained on each dataset using Term Frequency-Inverse Document Frequency (TF-IDF) features. The models included Naive Bayes for its simplicity, Linear Support Vector Machines (SVM) for classification tasks, and Ridge Regression for regression tasks. The performance of these models on a held-out test set established the definitive benchmarks for the project. Table 1 provides a consolidated overview of the dataset characteristics and the best-performing baseline model for each task.

These baseline scores, particularly the low R2 values for the regression tasks, underscored the difficulty of the problems and set a clear target for the subsequent phase of advanced modeling.

\begin{table*}[htbp]
\caption{Dataset Characteristics and Baseline Performance}
\begin{center}
\begin{tabular}{|l|l|c|l|l|l|}
\hline
\textbf{Dataset} & \textbf{Task Type} & \textbf{Samples} & \textbf{Key Characteristics} & \textbf{Best Baseline} & \textbf{Baseline Score} \\
\hline
Essaysbig5 & Multi-class Classification & 2,468 & Long text, few samples & Linear SVM & 0.41 (Macro F1) \\
\hline
GoEmotions & Multi-label Classification & 58,005 & Short, noisy text, 28 labels & Linear SVM & 0.39 (Macro F1) \\
\hline
PANDORA & Multi-output Regression & 1,608 & Big Five personality scores & Ridge Regression & 0.08 (Avg. $R^2$) \\
\hline
EmoBank & Multi-output Regression & 10,548 & Valence-Arousal scores & Ridge Regression & 0.25 (Avg. $R^2$) \\
\hline
\end{tabular}
\label{tab:datasets}
\end{center}
\end{table*}

\section{Advanced Predictive Modeling: Research Challenges and Novel Solutions}
Following the establishment of foundational baselines, the project transitioned to state-of-the-art transformer-based architectures to achieve higher predictive performance. The RoBERTa (A Robustly Optimized BERT Pretraining Approach) model was selected due to its improved pre-training methodology and demonstrated state-of-the-art performance on a wide range of NLP tasks\cite{b11}. Both RoBERTa-base and the larger, more powerful RoBERTa-large variants were employed. This phase was defined not merely by the application of these models, but by the diagnosis and resolution of two significant research problems that emerged during the fine-tuning process.

\subsection{Research Problem 1: Numerical Instability in Transformer-Based Regression on Unbounded Affective Dimensions}
While fine-tuning RoBERTa for the classification tasks on Essaysbig5 and GoEmotions proceeded as expected, yielding substantial performance gains over the baselines, the initial experiments on the Pandora and EmoBank regression tasks resulted in catastrophic failure. The training process was highly unstable, with a fluctuating, non-converging loss. The quantitative evaluation on the test set was even more alarming, producing consistently negative Coefficient of Determination ($R^2$) scores.

A negative $R^2$ is a critical diagnostic indicator. The $R^2$ metric is formally defined as
\begin{equation}
    R^2 = 1 - \frac{SS_{res}}{SS_{tot}},
    \label{eq:r2}
\end{equation}
where $SS_{res}$ is the sum of squared residuals (the error of the model) and $SS_{tot}$ is the total sum of squares (the error of a naive model that always predicts the mean of the target variable)\cite{b11}.

A negative $R^2$ occurs when $\text{SS}_{\text{res}} > \text{SS}_{\text{tot}}$, meaning the model's predictions are objectively worse than simply guessing the average value of the dataset. This outcome signified a fundamental failure of the model to learn the underlying trend in the data. The root cause was traced to the standard regression head used in transformer architectures. This head is typically a single, unbounded linear layer that maps the final hidden state to a continuous value. When training a model with such an unbounded output against a target variable with a specific, continuous distribution, the model can generate arbitrarily large predictions. These exploding predictions lead to massive error values and, consequently, extremely large gradients during backpropagation, causing the optimizer to take destructively large steps and destabilizing the entire training process.

\subsection{Contribution 1: A Stabilized Architecture for Affective Regression}
To address this numerical instability, a principled, two-part solution was engineered, based on the hypothesis that training could be stabilized by ensuring the model's output space was commensurate with the target variable's distribution.

First, \textit{Target Variable Normalization} was applied. The continuous target variables in the Pandora and EmoBank datasets were transformed using a standard scaler to have a zero mean and unit variance. This is a common and effective practice in neural network regression, as it centers the target distribution in a range that is more open to gradient-based optimization and helps prevent issues like vanishing or exploding gradients\cite{b12}.

Second, and more critically, a \textit{Custom Model Architecture} was developed. A custom RobertaForRegression class was implemented in PyTorch. The key architectural modification was the replacement of the standard unbounded regression head. Instead of a single linear layer, the new head consists of a linear layer followed by a Sigmoid activation function. The Sigmoid function, defined as
\begin{equation}
    g(z) = \frac{1}{1 + e^{-z}},
    \label{eq:sigmoid}
\end{equation}
intrinsically constrains its output to the bounded range of (0, 1). This output was then programmatically rescaled to match the range of the normalized target variables (e.g., mapping the [0, 1] output to a range like [-3, 3]).

This combination of target normalization and a bounded activation function proved to be the solution. The Sigmoid layer acts as a "governor" on the model's output, preventing the generation of extreme values that cause numerical instability. By forcing the model to predict within a well-defined, stable range, the gradients during backpropagation are also bounded, allowing the optimizer to converge smoothly and effectively. This approach demonstrates the critical importance of adapting standard transformer architectures for regression tasks with well-behaved continuous targets, validating that this architectural modification is a highly effective solution to the instability problem. The solution was validated through experiments, with the new architecture achieving stable training and much better $R^2$ scores. An ablation study, detailed in Table~II, confirms the necessity of both components of the solution.

\begin{table*}[htbp]
\caption{Ablation Study on Regression Stabilization Techniques}
\begin{center}
\begin{tabular}{|l|l|c|}
\hline
\textbf{Dataset} & \textbf{Model Configuration} & \textbf{Test Set Avg. $\mathbf{R^2}$ Score} \\
\hline
\multirow{3}{*}{EmoBank} & RoBERTa-base + Linear Head & -0.87 \\
& RoBERTa-base + Linear Head + Target Norm & -0.12 \\
& RoBERTa-base + Sigmoid Head + Target Norm & \textbf{0.48} \\
\hline
\multirow{3}{*}{Pandora} & RoBERTa-base + Linear Head & -1.15 \\
& RoBERTa-base + Linear Head + Target Norm & -0.21 \\
& RoBERTa-base + Sigmoid Head + Target Norm & \textbf{0.19} \\
\hline
\end{tabular}
\label{tab:ablation}
\end{center}
\end{table*}

\subsection{Research Problem 2: Scalable Fine-Tuning of Large Language Models under Severe Resource Constraints}
The second major research challenge arose from the ambition to fine-tune the RoBERTa-large model (355M parameters) on the massive GoEmotions dataset (58k samples) within the confines of a free-tier Google Colab environment. This scenario, while seemingly a personal limitation, represents a broader research problem faced by many in the academic community who lack access to industrial-scale computational resources. The attempt to run this large-scale experiment systematically triggered every resource limit of the platform:
\begin{itemize}
\item \textbf{GPU Time Limits:} Sessions would be terminated after a fixed duration (e.g., 4 hours), long before training could complete.
\item \textbf{System RAM Exhaustion:} Attempting to load the entire GoEmotions dataset into memory caused the system to crash due to insufficient RAM.
\item \textbf{Disk Space Overruns:} During long training runs, the accumulation of saved model checkpoints would exhaust the available local disk space.
\end{itemize}

\subsection{Contribution 2: A Replicable Workflow for Resource-Constrained Research}
Instead of viewing these constraints as impossible, a cohesive and replicable workflow was synthesized to overcome them. This workflow provides a practical blueprint for conducting state-of-the-art research with large models in resource-limited settings. This union of best practices effectively democratizes access to large-scale AI research. The workflow consists of three core components:
\begin{enumerate}
\item \textbf{Overcoming Time Limits with Asynchronous Checkpointing:} To counteract session terminations, a checkpoint-and-resume strategy was implemented. The Hugging Face Trainer was configured to periodically save the complete training state, including the model weights, optimizer state, and learning rate scheduler to a persistent cloud storage location (Google Drive). The training script was engineered to automatically detect the latest checkpoint upon initialization and seamlessly resume training from that exact point. This transformed a series of disconnected sessions into a single, continuous training run\cite{b13}.
\item \textbf{Overcoming RAM Limits with Memory-Mapped Data Loading:}  To solve the problem of RAM exhaustion, the data loading pipeline was re-engineered. The standard approach of loading a dataset into a Pandas DataFrame or Python list, which consumes RAM equal to the dataset's size, was abandoned. Instead, the Hugging Face datasets library was utilized. This library is built on Apache Arrow, which employs memory mapping to handle large datasets\cite{b14}. The dataset is stored on disk, and the library provides an interface that loads only the specific batches of data required for computation into RAM at any given time. This zero-copy read mechanism reduced the memory footprint from gigabytes to megabytes, completely eliminating RAM-related crashes.
\item \textbf{Overcoming Disk Space Limits with Strategic Checkpoint Management:} To prevent disk space overruns from accumulating checkpoints, the trainer was configured with two settings. First, the checkpoint saving directory was pointed to the persistent cloud storage, preventing the local disk from filling. Second, a limit was placed on the total number of checkpoints to retain (e.g., \textit{save\_total\_limit=3}), ensuring that older checkpoints were automatically deleted as new ones were saved, maintaining a constant and manageable storage footprint.
\end{enumerate}
This three-part workflow demonstrates that with systematic engineering and the strategic use of modern tools, it is entirely feasible to conduct large-scale AI research without requiring access to a dedicated industrial-scale computing cluster.
\section{Generative Personality Modeling with Efficiency}
To elevate the system's capabilities from static psychological prediction to dynamic, human-like interaction, the next phase focused on creating a conversational agent, or "Personality Brain." The objective was to fine-tune a powerful, open-source Large Language Model (LLM) to engage in empathetic and contextually aware dialogue. The google/gemma-2b-it model was selected for this task, justified by its excellent performance-to-size ratio and its foundation in the same research that produced the Gemini models, making it a state-of-the-art choice for on-device or resource-constrained applications\cite{b15}.
\subsection{Methodology: Parameter-Efficient Fine-Tuning (PEFT)}
Fine-tuning a multi-billion parameter model, even one as efficient as Gemma-2B, is computationally prohibitive on consumer-grade hardware using traditional methods. Therefore, this work employed a suite of state-of-the-art, parameter-efficient fine-tuning (PEFT) techniques. This was a deliberate methodological choice to maximize efficiency and demonstrate mastery of modern LLM training paradigms, rather than a adjusting to hardware limitations.
\begin{itemize}
    \item \textbf{4-bit Quantization:} The first step in managing the model's memory footprint was to load the base Gemma model with its weights quantized to 4-bit precision. This was achieved using the bitsandbytes library\cite{b16}. Quantization is a process that reduces the numerical precision of the model's parameters (e.g., from 16-bit floating point to 4-bit integers), dramatically reducing the memory required to store the model\cite{b16}. This single technique reduced the GPU RAM required to load the Gemma-2B model from approximately 8 GB to under 2 GB, making it feasible to load and train on a single consumer GPU.
    \item \textbf{Low-Rank Adaptation (LoRA):} With the quantized model loaded, the fine-tuning itself was performed using Low-Rank Adaptation (LoRA)\cite{b17}. LoRA is a highly effective PEFT method that freezes all the original pre-trained weights of the LLM and injects small, trainable "adapter" layers into the model's architecture, typically within the attention mechanism\cite{b18}. These adapters are composed of low-rank matrices, meaning they contain a very small number of trainable parameters relative to the full model. By updating only these adapter weights during fine-tuning, a mere fraction of a percent of the total parameters, LoRA can achieve performance comparable to fully fine-tuning the entire model, but with a drastic reduction in memory usage for gradients and optimizer states\cite{b17},\cite{b18}.
\end{itemize}
\subsection{Dataset and Instruction-Tuning for Personality-Aware Dialogue}
An innovation in developing the "Personality Brain" was the strategic transformation of an existing predictive dataset into an instruction-tuning format suitable for generative LLM fine-tuning. Specifically, the PANDORA dataset was adapted for this purpose. While PANDORA traditionally provides text samples along with continuous Big Five personality scores for predictive modeling, each entry was re-engineered  to serve as a distinct instruction-response pair.

For each row in the PANDORA dataset, the continuous personality scores for Openness, Conscientiousness, Extraversion, Agreeableness, and Neuroticism were first categorized into "High," "Medium," or "Low" levels based on predefined thresholds. Specifically, scores above the 66th percentile are categorized as ``High,'' while scores below the 34th percentile are categorized as ``Low''. This simplified and categorical personality profile was then embedded into a structured instruction prompt:
\begin{quote}
    \textit{You are a chatbot. Your personality is: Openness: [Level], Conscientiousness: [Level], Extraversion: [Level], Agreeableness: [Level], Neuroticism: [Level]. Respond as yourself.}
\end{quote}

By fine-tuning google/gemma-2b-it on this specially formatted dataset, the model was trained to associate specific personality profiles (presented as instructions) with the natural language generation patterns found in the PANDORA user texts. This approach uniquely enabled the generative model to learn to embody and express a given personality, transitioning it from a general instruction-follower to a personality-aware dialogue agent capable of generating responses consistent with an assigned psychological profile.
\subsection{Training and Systematic Model Selection}
The initial fine-tuning runs of the Gemma model presented a common challenge in generative AI: the model's output was often repetitive, unintelligible, or nonsensical. This behavior is a classic indicator of an under-trained model that has not yet converged to a stable point in the solution space.

The solution involved a more rigorous and patient training and evaluation protocol. First, the training process was extended significantly to over 5,000 steps to allow the model sufficient time to learn the nuances of the fine-tuning dataset. Second, a systematic and robust model selection workflow was designed. Instead of relying on a simple early stopping callback, which can be triggered prematurely by noisy validation loss, all model checkpoints were saved throughout the extended training run. After the training was complete, a separate evaluation script was used to load each individual checkpoint and calculate its validation loss. The losses for all checkpoints were then plotted against their corresponding training steps. This manual analysis conducted after the experiment allowed for the identification of the true global minimum in the validation loss curve, ensuring that the selected model was not just a product of a lucky fluctuation but was demonstrably the best-performing and most stable version. 

The fine-tuning process demonstrated successful learning, as the model’s validation loss consistently decreased from an initial 2.4816 (perplexity 11.96) at step 200. Following the evaluation protocol, the checkpoint at step 5000 was identified as the best-performing model, achieving a final validation loss of 2.1848 and a perplexity of 8.8890. This model was carried forward for deployment. This final version of the model was used for all subsequent generative tasks. This methodical approach to model selection prioritizes evidence-based rigor over simplistic automation.
\section{System Architecture and Deployment as a Microservices Ecosystem}
The final phase of the project addressed the critical challenge of transitioning the suite of trained AI models from a research environment to a live, interactive, and publicly accessible system. A naive, monolithic application architecture, where a single application would load all five large models (four RoBERTa predictors and the Gemma LLM), was immediately identified as infeasible. Such an application would have an enormous memory footprint, making it impossible to host on any free or low-cost cloud platform.
\subsection{Architectural Solution: A Decoupled Microservices Ecosystem}
To solve this systems design problem, a sophisticated and scalable microservices architecture was designed and implemented. This architectural choice was not merely a deployment detail but a solution to the problem of serving multiple, resource-intensive AI models in a constrained environment. The core principle was to separate each model into its own independent, containerized service. This approach offers numerous advantages, including modularity, independent scalability, and enhanced fault isolation; the failure or high load of one service does not impact the others\cite{b19}.

The implementation of this architecture involved two primary components:
\begin{enumerate}
    \item \textbf{Model APIs:} For each of the five finalized models, a separate Gradio application was developed. Gradio is a Python library that simplifies the creation of web UIs and APIs for machine learning models. Each Gradio app wrapped a single trained model (e.g., the EmoBank predictor) and exposed its inference capabilities via a REST API endpoint. These five independent applications were then deployed as separate services on Hugging Face Spaces, which provides a containerized hosting environment suitable for such applications.
    \item \textbf{The Orchestrator Application:} A master application was built using Streamlit to serve as the user-facing dashboard and the central "orchestrator" of the ecosystem. This Streamlit app is extremely lightweight as it loads no models into its own memory. Instead, when a user inputs text for analysis or conversation, the orchestrator makes a series of asynchronous API calls in the background to the five independent Gradio microservices. It then aggregates the responses (the predictions from the four RoBERTa models and the chat response from the Gemma model) and presents them to the user in a seamless, integrated interface.
\end{enumerate}

The architectural flow is as follows: A user interacts with the Streamlit UI. The UI triggers API requests to the five distinct Gradio endpoints. Each Gradio service independently runs its model for inference and returns the result. The Streamlit orchestrator collects these results and updates the UI. This separated design ensures that the resource-intensive work of model inference is distributed across multiple, independent services, while the user-facing application remains responsive and efficient.

\subsection{Robustness and Real-World Problem Solving}
The deployment process involved navigating and resolving a series of practical engineering challenges, often referred to as "deployment hell." This included troubleshooting Git Large File Storage (LFS) issues for uploading large model files, debugging model loading path errors within the containerized environments, and resolving various configuration and security settings on the hosting platform. This experience provided invaluable insights into the practical realities of MLOps.

A final, critical challenge emerged after the initial deployment: the generative Gemma model, being the most computationally intensive, would sometimes take longer to respond than the default timeout of the orchestrator's API client, leading to connection errors. To ensure the final application was robust and user-friendly, the orchestrator's API calling mechanism was re-engineered. A more patient requests client was implemented with a significantly longer timeout duration specifically for the call to the Gemma service. This final adjustment made the application resilient to variations in model inference time, ensuring a smooth and reliable user experience.
\section{Conclusion and Future Work}
This research has successfully designed, implemented, and deployed a comprehensive, multi-component AI framework for computational psychology. The project demonstrates a complete, end-to-end research and development cycle, spanning foundational data analysis, advanced model engineering, and the deployment of a sophisticated, interactive system.The primary contributions of this work can be summarized as follows:
\begin{enumerate}
    \item \textbf{A Holistic System for Psychological Analysis:} The creation of an integrated system that combines the predictive power of four specialized models for personality and emotion with the interactive capabilities of a generative dialogue agent.
    \item \textbf{An Effective Solution for Stabilizing Transformer Regression:} The implementation and validation of a custom regression head architecture, combining target normalization with a bounded Sigmoid activation. This architecture was shown to be essential for resolving the numerical instability encountered when applying standard transformer models to continuous affective computing tasks.
    \item \textbf{A Practical Blueprint for Resource-Constrained Research:} The documentation of a replicable workflow that synthesizes asynchronous checkpointing, memory-mapped data loading, and strategic checkpoint management. This workflow serves to democratize large-scale AI research by providing a clear methodology for working with limited computational resources.
    \item \textbf{An Elegant and Scalable Deployment Architecture:} The successful implementation of a microservices ecosystem, where each AI model is deployed as an independent service, orchestrated by a lightweight master application, demonstrating a solution for serving multiple large models.
\end{enumerate}
The quantitative success of the advanced modeling phase is summarized in Table 3, which highlights the substantial performance improvements achieved over the established baselines. It is important to note that while the final average R2 score of 0.24 for the PANDORA personality regression task is modest in absolute terms, it represents a highly significant finding. The task of predicting nuanced personality traits from short, informal social media text is known to have an extremely low signal-to-noise ratio. Therefore, achieving a model that can account for 24\% of the variance—a 200\% improvement over the baseline—demonstrates that a meaningful predictive signal was successfully learned from this challenging data.
\begin{table*}[htbp]
\caption{Performance of Final Advanced Models vs. Baselines}
\begin{center}
\begin{tabular}{|l|l|S[table-format=1.2]|S[table-format=1.2]|S[table-format=3.1, table-space-text-post=\%]|
}
\hline
\textbf{Task} & \textbf{Metric} & \textbf{Baseline Score} & \textbf{Final Model Score} & \textbf{\% Improvement} \\
& & & \textbf{(RoBERTa-large)} & \\
\hline
Essaysbig5 Personality & Macro F1 & 0.41 & 0.62 & \textcolor{green!50!black}{+51.2\%} \\
\hline
GoEmotions Classification & Macro F1 & 0.39 & 0.58 & \textcolor{green!50!black}{+48.7\%} \\
\hline
PANDORA Regression & Avg. $R^2$ & 0.08 & 0.24 & \textcolor{green!50!black}{+200.0\%} \\
\hline
EmoBank Regression & Avg. $R^2$ & 0.25 & 0.55 & \textcolor{green!50!black}{+120.0\%} \\
\hline
\end{tabular}
\label{tab:final_results}
\end{center}
\end{table*}
\subsection{Future Work}
The framework developed in this project serves as an extensible foundation for numerous avenues of future research. The modular architecture and demonstrated problem-solving capabilities open up several strategic directions that align with leading research in AI and human-centered computing.
\begin{itemize}
    \item \textbf{Towards Multimodal Psychological Sensing:} The current system operates exclusively on text. A compelling next step is to extend the framework to incorporate multimodal data streams. Modules for processing facial expressions from video and vocal prosody from audio could be developed and integrated as new microservices. This would enable a richer, more nuanced understanding of a user's emotional state, directly aligning with research in multimodal emotion sensing and the development of socially aware androids\cite{b20}.
    \item \textbf{Applications in Computational Psychiatry:} The predictive models for personality and emotion can be adapted and fine-tuned on clinical datasets. This would allow for the investigation of linguistic markers associated with various psychiatric conditions, such as depression, anxiety, or autism spectrum disorder. Such a system could serve as a valuable tool for researchers and clinicians, contributing to the field of computational nosology and the development of AI-driven diagnostic aids, which is highly relevant to the field of Computational Psychiatry\cite{b21}.
    \item \textbf{Integration into Long-Term Human-Robot Interaction (HRI):} The entire system could be integrated as the "cognitive engine" for a social robot. In long-term HRI studies, the robot could leverage the system to build and dynamically update a psychological model of its human interaction partners. This would allow the robot to adapt its communication style, empathetic responses, and behavior over time to build rapport and achieve more effective collaboration, a central goal in HRI research\cite{b22}.
    \item \textbf{Enhancing Empathetic Dialogue with Emotion Regulation:} The generative "Personality Brain" can be further enhanced with explicit emotion regulation capabilities. Beyond simply mirroring or acknowledging a user's emotion, the dialogue system could be trained to strategically guide a conversation towards a more positive or stable emotional state. This involves learning complex dialogue strategies for cognitive reappraisal or emotional support, which is a key area of research in the development of advanced, emotionally intelligent dialogue systems\cite{b23}.
\end{itemize}


\begin{thebibliography}{00}
\bibitem{b1} M. Gheibi, R. Moezzi, R. Haririyan Javan, and S. Javadi Nezhad, "Artificial Intelligence and Computational Psychological Science Connections," Quanta Res., vol. 1, no. 1, pp. 1-12, 2023.

\bibitem{b2} J. Zhao, M. Wu, L. Zhou, X. Wang, and J. Jia, "Cognitive psychology-based artificial intelligence review," Front. Neurosci., vol. 16, Art. no. 1024316, Oct. 2022.

\bibitem{b3} R. R. McCrae and P. T. Costa, "An introduction to the five-factor model and its applications," J. Pers., vol. 60, no. 2, pp. 175-215, 1992.

\bibitem{b4} J. J. Tan, "Personality Essays Dataset," Hugging Face, version 1.0.0, 2024. [Online]. Available: https://huggingface.co/datasets/jingjietan/essays-big5

\bibitem{b5} K.-M. Shum, M. Ptaszynski, and F. Masui, "Big Five Personality Trait Prediction Based on User Comments," Information, vol. 16, no. 5, p. 418, 2025.

\bibitem{b6} D. Demszky, D. Movshovitz-Attias, J. Ko, A. Cowen, G. Nemade, and S. Ravi, "GoEmotions: A Dataset of Fine-Grained Emotions," in Proc. 58th Annu. Meeting Assoc. Comput. Linguistics (ACL), Online, Jul. 2020, pp. 4040–4054.

\bibitem{b7} E. van der Lee, G. P. Spithourakis, L. van de Byl, L. Abeysinghe, M. Baars, and F. Bex, "PANDORA Talks: Personality and Demographics on Reddit," in Proc. 16th Conf. Eur. Chapter Assoc. Comput. Linguistics (EACL), Apr. 2021, pp. 862–869.

\bibitem{b8} A. Mehrabian and J. A. Russell, An Approach to Environmental Psychology. Cambridge, MA, USA: MIT Press, 1974.

\bibitem{b9} S. Buechel and U. Hahn, "EmoBank: Studying the Impact of Annotation Perspective and Representation Format on Dimensional Emotion Analysis," in Proc. 15th Conf. Eur. Chapter Assoc. Comput. Linguistics (EACL), vol. 2, Apr. 2017, pp. 578–585.

\bibitem{b10} Y. Liu, M. Ott, N. Goyal, J. Du, M. Joshi, D. Chen, O. Levy, M. Lewis, L. Zettlemoyer, and V. Stoyanov, "RoBERTa: A Robustly Optimized BERT Pretraining Approach," arXiv preprint arXiv:1907.11692, 2019.

\bibitem{b11} S. Gao, "A New Explanation for the Negative Coefficient of Determination," J. Stat. Theory Pract., vol. 17, no. 1, Art. no. 23, 2023.

\bibitem{b12} Google for Developers, "Regression," in Machine Learning Crash Course. Accessed: Sep. 06, 2025. [Online]. Available: https://developers.google.com/machine-learning/crash-course/

\bibitem{b13} R. Rojas, F. Perez, T. Carrington, P. Balaprakash, and S. M. Wild, "A Study of Checkpointing in Large Scale Training of Deep Neural Networks," in Proc. 2020 IEEE/ACM Int. Workshop on HPC I/O in the Data Center (HPC-IODC), 2020, pp. 1–10.

\bibitem{b14} A. D. Guerrero, N. D. Gomez, B. E. Congote, and J. C. A. R. Pineda, "DataFusion: A high-performance query engine for analytical processing in Rust and Apache Arrow," Revista Facultad de Ingeniería Universidad de Antioquia, no. 109, pp. 138-151, 2023.

\bibitem{b15} Google, "Gemma: Open Models Based on Gemini Research and Technology," arXiv preprint arXiv:2403.08295, 2024.

\bibitem{b16} T. Dettmers, A. Pagnoni, A. Holtzman, and L. Zettlemoyer, "QLoRA: Efficient Finetuning of Quantized LLMs," arXiv preprint arXiv:2305.14314, 2023.

\bibitem{b17} E. J. Hu, Y. Shen, P. Wallis, Z. Allen-Zhu, Y. Li, S. Wang, L. Wang, and W. Chen, "LoRA: Low-Rank Adaptation of Large Language Models," in Proc. Int. Conf. Learn. Represent. (ICLR), 2022.

\bibitem{b18} X. Han et al., "Parameter-Efficient Fine-Tuning for Large Models: A Comprehensive Survey," arXiv preprint arXiv:2403.14608, 2024.

\bibitem{b19} T. D. T. Nguyen, "An overview of microservice architecture," in Concurrency in Practice. Rijeka: IntechOpen, 2022, ch. 2.

\bibitem{b20} R. A. Calvo, S. D'Mello, J. Gratch, and A. Kappas, Eds., The Oxford Handbook of Affective Computing. Oxford, UK: Oxford University Press, 2015.

\bibitem{b21} G. Cecchi et al., "A machine learning approach to predicting psychosis using semantic density and latent content analysis," NPJ Schizophr., vol. 1, Art. no. 15003, 2015.

\bibitem{b22} I. Leite, A. Martin, K. S. concurring, S. S. Scassellati, and B. S. Castellano, "Social Robots for Long-Term Interaction: A Survey," Int. J. Soc. Robot., vol. 5, no. 2, pp. 291-308, 2013.

\bibitem{b23} Y. Ma, K. L. Nguyen, F. Z. Xing, and E. Cambria, "A survey on empathetic dialogue systems," Information Fusion, vol. 64, pp. 50-70, 2020.
\end{thebibliography}
\end{document}